\documentclass{article} 
\usepackage{iclr2017_workshop,times}
\usepackage{hyperref}
\hypersetup{
    colorlinks = true,
    linkcolor = blue,
    linkbordercolor = {white},
    citecolor=blue,
}
\usepackage{url}
\usepackage{todonotes}

\usepackage{amsmath}
\usepackage{amsfonts}
\usepackage{amsthm}
\usepackage{graphicx} 
\usepackage{caption}
\usepackage{subcaption} 
\usepackage{float}

\usepackage[capitalise]{cleveref}
\newcommand{\ie}[0]{\emph{i.e.},~}
\newcommand{\eg}[0]{\emph{e.g.},~}
\newcommand{\aka}[0]{a.k.a.~}

\newcommand{\yy}[0]{\ensuremath{\mathbf{y}}}
\newcommand{\xx}[0]{\ensuremath{\mathbf{x}}}
\newcommand{\ff}[0]{\ensuremath{\mathbf{f}}}

\newcommand{\YY}[0]{\ensuremath{\mathcal{Y}}}
\newcommand{\XX}[0]{\ensuremath{\mathcal{X}}}

\newcommand{\comment}[1]{{}}

\newtheorem{theorem}{Theorem}[section]

\usepackage{xcolor}
\usepackage[framemethod=TikZ]{mdframed}

\mdfdefinestyle{MyFrame}{%
    linecolor=white,
    outerlinewidth=1pt,
    roundcorner=10pt,
    innertopmargin=\baselineskip,
    innerbottommargin=\baselineskip,
    innerrightmargin=5pt,
    innerleftmargin=5pt,
    backgroundcolor=gray!20!white}

\title{Deep Learning with Sets and Point Clouds}

\author{Siamak Ravanbakhsh, Jeff Schneider \& Barnab{\'{a}}s P{\'{o}}czos\\
School of Computer Science\\
Carnegie Mellon University\\
Pittsburgh, PA 15213, USA \\
\texttt{\{mravanba,jeff.schneider,bapoczos\}@cs.cmu.edu} \\
}

\begin{document}

\maketitle
\begin{abstract}
We introduce a simple permutation equivariant layer for deep learning with set structure.
This type of layer, obtained by parameter-sharing, has a simple implementation and linear-time complexity in the size of each set.
We use deep permutation-invariant networks to perform
point-could classification and MNIST-digit summation, where in both cases the output is invariant to permutations of the input.
In a semi-supervised setting, where the goal is make predictions for each instance within a set,
we demonstrate the usefulness of this type of layer 
in set-outlier detection as well as semi-supervised learning with clustering side-information.
\end{abstract}

\section{Introduction}

Recent progress in deep learning~\citep{lecun2015deep} has witnessed its application to structured settings, including graphs~\citep{bruna2013spectral,duvenaud2015convolutional} groups \citep{gens2014deep,gconv,cohen2016group}, sequences and hierarchies~\citep{irsoy2014deep,socher2013recursive}.
Here, we introduce a simple permutation-equivariant layer for deep learning with set structure, where the primary dataset is a collection of sets, possibly of different sizes. 
Note that each instance may have a structure of its own, such as graph, image, or another set. 
In typical machine-learning applications, iid assumption implies that
the entire data-set itself has a set structure. Therefore, our special treatment of the set structure is only necessary due to multiplicity of distinct yet homogeneous (data)sets. 
Here, we show that a simple parameter-sharing scheme enables a general treatment of sets within supervised and semi-supervised settings.

In the following, after introducing the set layer in \cref{sec:parameter_sharing}, we explore several novel applications. 
\cref{sec:supervised} studies supervised learning with sets that requires ``invariance'' to permutation of inputs. 
\Cref{sec:mnist} considers the task of summing multiple MNIST digits, and \cref{sec:pointcloud} studies an important application of sets in representing low-dimensional point-clouds. 
Here, we show that deep networks can successfully classify objects using their point-cloud representation.

\cref{sec:semi_supervised} presents numerical study in semi-supervised setting where the output of the multi-layer network is ``equivariant'' to the permutation of inputs.
We use permutation-equivariant layer to perform outlier detection on CelebA face dataset in \cref{sec:celeb} and 
improve galaxy red-shift estimates using its clustering information in \cref{sec:galaxy}.

\section{Permutation-Equivariant Layer}\label{sec:parameter_sharing}
Let $x_n \in \XX$ denote use $\xx = [x_1,\ldots,x_N]$ to denote a vector of $x_n$ instances.
Here, $x_n$, could be a feature-vector, an image or any other structured object. Our goal is to design 
neural network layers that are ``indifferent'' to permutations of instances in $\xx$. Achieving this goal
amounts to treating $\xx$ as a ``set'' rather than a vector.

The function $\ff: \XX^{N} \to \YY^{N}$ is \textbf{equivariant} to the permutation of its inputs iff 
\begin{align*}
  \ff(\pi \xx) = \pi \ff(\xx) \quad \forall \pi \in \mathcal{S}_N
\end{align*}
where the symmetric group $\mathcal{S}_N$ is the set of all permutation of indices $1,\ldots,N$.
Similarly, the function $f: \Re^{N} \to \Re$ is \textbf{invariant} to permutation of its inputs --\ie \aka a symmetric function~\citep{david1966symmetric}--
iff $f(\pi \xx) = f(\xx)\quad \forall \pi \in \mathcal{S}_N$.
Here, the \textbf{action} of $\pi$ on a vector $\xx \in \Re^{N}$ can be \textit{represented} by a permutation matrix. 
With some abuse of notation, we use $\pi \in \{0,1\}^{N \times N}$ to also denote this matrix.

Given two permutation equivariance function $\ff: \XX^{N} \to \YY^{N}$ and $\mathbf{g}: \YY^{N} \to \mathcal{Z}^{N}$,
their \textbf{composition} is also permutation-equivariance; this is because $\mathbf{g}(\ff(\pi \xx)) = \mathbf{g}(\pi \ff(\xx)) = \pi \mathbf{g}(\ff(\xx))$.

Consider the standard neural network layer
\begin{align}\label{eq:1}
  \ff_{\Theta}(\xx) \doteq \boldsymbol{\sigma}( \Theta \xx) \quad \Theta \in \Re^{N \times N}
\end{align}
where $\Theta$ is the weight vector and $\sigma: \Re \to \Re$ is a nonlinearity such as sigmoid function. $\boldsymbol{\sigma}:\Re^{N} \to \Re^{N}$ is the point-wise application of $\sigma$ to its input vector. The following theorem states the necessary and sufficient conditions for permutation-equivariance in this type of function. 

\begin{theorem}\label{theorem}
  The function $\ff_{\Theta}: \Re^{N} \to \Re^{N}$ as defined in \cref{eq:1} is permutation equivariant
iff all the off-diagonal elements of $\Theta$ are tied together and all the diagonal elements are equal as well,
\begin{align}
  \label{eq:2}
 \Theta = {\lambda} \mathbf{I} + {\gamma} \; (\mathbf{1} \mathbf{1}^{\mathsf{T}})\quad\quad {\lambda},{\gamma} \in \Re \quad \mathbf{1} = [1,\ldots,1]^{\mathsf{T}} \in \Re^N
\end{align}
where $\mathbf{I}$ is the identity matrix.\footnote{See the \cref{sec:proof} for the proof.}
\end{theorem}
This function is simply a non-linearity applied to a weighted combination of I) its input $ \mathbf{I} \xx$ and; II) the sum of input values $(\mathbf{1} \mathbf{1}^{\mathsf{T}}) \xx$. 
Since summation does not depend on the permutation, the layer is 
permutation-equivariant. 
Therefore we can manipulate the operations and parameters in this layer, for example to get another \textbf{variation}
\begin{align}
  \label{eq:3}
 \ff(\xx) \doteq \boldsymbol{\sigma}(\lambda \mathbf{I} \xx - \gamma (\max_{n} \xx) \mathbf{1}) 
\end{align}
where the max operation over elements of the set is (similar to summation) commutative and using $-\gamma$ instead of $+\gamma$ amounts to a reparametrization.
In practice using this variation performs better in some applications. This may be due to the fact that for $\lambda = \gamma$, the input to the non-linearity is max-normalized.

\begin{figure}[t]
  \centering
\includegraphics[width=\textwidth]{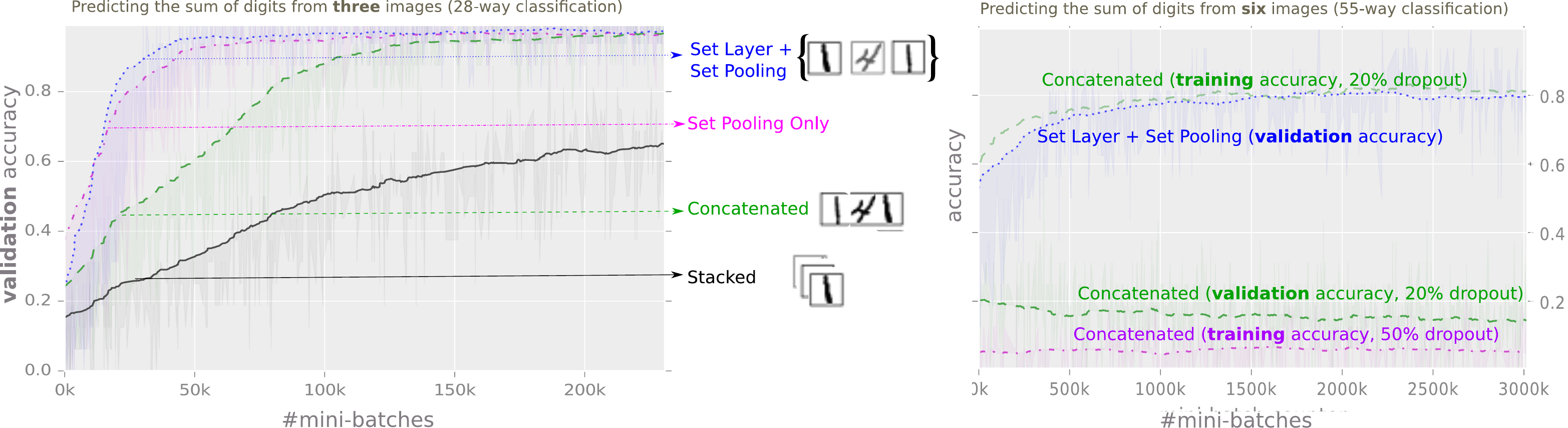} 
  \caption{\it \small Classification accuracy of different schemes in predicting the sum of a (left) N=3 and (right) N=6 MNIST digits without access to individual
image labels. The training set was fixed to 10,000 sets.} \label{fig:mnist}
\end{figure}

For \textbf{multiple input-output channels}, we may speed up the operation of the layer using matrix multiplication.
Suppose we have $K$ input channels --corresponding to $K$ features for each instance in the set-- with a set of size $N$, and $K'$ output channels. Here, $\xx \in \Re^{N \times K}$ and $\ff_{\Theta}: \Re^{N \times K} \to \Re^{N \times K'}$.
The permutation-equivariant layer parameters are ${\Lambda}, \Gamma \in \Re^{K,K'}$ (replacing $\lambda$ and $\gamma$ in \cref{eq:2}). The output of this layer becomes
\begin{align*}
  \yy = \boldsymbol{\sigma} \big(
\xx \Lambda \, +\,  \mathbf{1} \,\mathbf{1}^{\mathsf{T}} \xx \Gamma
\big)
\end{align*}
Similarly, the multiple input-output channel variation of the \cref{eq:3} is
\begin{align}
  \label{eq:4}
 \yy = \boldsymbol{\sigma} \big(
\xx \Lambda \, -\,  \mathbf{1} \xx_{\max} \Gamma
\big) 
\end{align}
where $\xx_{\max} = (\max_n \xx) \in \Re^{1 \times K}$ is a row-vector of maximum value of $\xx \in \Re^{N \times K}$ over the ``set'' dimension.
In practice, we may further reduce the number of parameters in favor of better generalization by factoring $\Gamma$ and $\Lambda$ and keeping
a single $\lambda \in \Re^{K,K'}$
\begin{align}
  \label{eq:5}
 \yy = \boldsymbol{\sigma} \big( \beta + 
\big (\xx \, -\,  \mathbf{1} (\max_{n} \xx) \big ) \Gamma
\big) 
\end{align}
where $\beta \in \Re^{K'}$ is a bias parameter. This final variation of permutation-equivariant layer
is simply a fully connected layer where input features are max-normalized within each set.

With multiple input-output channels, the complexity of this layer for each set is $\mathcal{O}(N\ K\ K')$. 
Subtracting the mean or max over the set also reduces the internal covariate shift~\citep{ioffe2015batch} 
and we observe that for deep networks (even using tanh activation), batch-normalization is not required.

When applying dropout~\citep{srivastava2014dropout} to regularize permutation-equivariant layers with multiple output channels,
it is often beneficial to \textit{simultaneously} \textbf{dropout} the channels for all instances within a set.
In particular, when set-members share similar features, independent dropout effectively does not regularize the 
model as the the network learns to replace the missing features from other set-members.

In the remainder of this paper, we demonstrate how this simple treatment of sets can solve
novel and non-trivial problems that occasionally have no alternative working solutions within deep learning.

\section{Supervised Learning}\label{sec:supervised}
The permutation-equivariant layers that we introduced so far are useful for
semi-supervised learning (or \textit{transductive}) setting, where we intend to predict a value per each instance in every set.
In supervised (or \textit{inductive}) setting the task is to make a prediction for each set (rather than instances within them), and we require permutation ``invariance'' of $f:\XX^{N} \to \YY$.
A pooling operation over the set-dimension can turn any permutation equivariant function $\ff: \XX^{N} \to \YY^{N}$, permutation invariant $f(\xx) \doteq \bigoplus_{n} \ff(\xx)$. Here $\oplus$ is any commutative operation such as summation or maximization. 

In a related work \cite{chen2014unsupervised} construct deep permutation invariant features by 
pairwise coupling of features at the previous layer, where $\ff_{i,j}([x_i, x_j]) \doteq [|x_i - x_j|, x_i+x_j]$ 
is invariant to transposition of $i$ and $j$.\footnote{Due to change in the number of features/channels in each layer this approach cannot produce
permutation-``equivariant'' layers. 
Also, this method requires a graph to guide the multi-resolution partitioning of the nodes, which is then used to define pairing of features in each layer.}
In another related work \citet{vinyals2015order} approach unordered instances by finding ``good'' orderings.

As we see shortly in \cref{sec:mnist}, in the supervised setting, even simple application of set-pooling, without 
max-normalization of \cref{eq:5} performs very well in practice, where we need permutation invariance.
However, in semi-supervised setting
since there is no pooling operation, the permutation invariant layer requires max-normalization
in order to obtain the required information about the \textit{context} of each instance -- \ie permutation equivariance.

\begin{figure}
  \centering
\includegraphics[width=1\textwidth,trim={0 5em 0 0},clip]{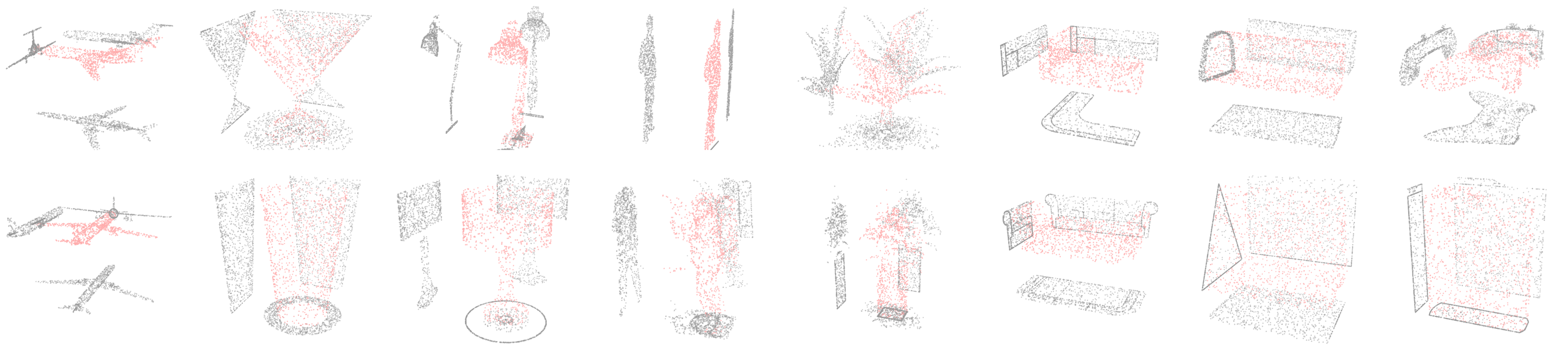} 
  \caption{\it \small Examples for 8 out of 40 object classes (column) in the ModelNet40. Each point-cloud is produces by sampling 1000 particles from the mesh representation of the original MeodelNet40 instances. Two point-clouds in the same column are from the same class. The projection of particles into xy, zy and xz planes are added for better visualization. 
} \label{fig:modelnet_dataset}
\end{figure}

\subsection{Predicting the Sum of MNIST Digits}\label{sec:mnist}
MNIST dataset~\citep{lecun1998gradient} contains 70,000 instances of $28 \times 28$ grey-scale stamps of digits in $\{0,\ldots,9\}$.
We randomly sample a subset of $N$ images from this dataset to build 10,000 ``sets'' of training and 10,000 sets of validation images, 
where the set-label is the sum of digits in that set (\ie individual labels per image is unavailable).

In our first experiment, each set contains $N=3$ images, and the set label is a number between $0\leq y \leq 3*9 = 27$.
We then considered four different models for predicting the sum:\\\noindent

\textbf{I}) Concatenating instances along the horizontal axis.\\\noindent
\textbf{II}) Stacking images in each set as different input channels.\\\noindent
\textbf{III}) Using set-pooling without max-normalization.\\\noindent
\textbf{IV}) Using the permutation equivariant layer of \cref{eq:5} with set-pooling.\\\noindent

All models are defined to have similar number of layers and parameters; see \cref{sec:mnist_details} for details.
The output of all models is a ($9 N+1$)-way softmax, predicting the sum of $N$ digits. 

\Cref{fig:mnist}(left) show the prediction accuracy over the validation-set for different models for $N=3$.
We see that using the set-layer performs the best. However, interestingly, using
set-pooling alone produces similarly good results. 
We also observe that concatenating the digits eventually performs well, despite its lack of invariance. 
This is because due to the sufficiently large size of the dataset most permutations of length three appear in the training set.

However, as we increase the size of each set to $N = 6$, permutation invariance becomes crucial; see  \Cref{fig:mnist}(right).
We see that using the default dropout rate of $20\%$, the model simply memorizes the input instances (indicated by discrepancy of training/validation error) 
and by increasing this dropout rate, the model simply predicts values close to the mean value. However, the permutation-invariant layer
learns to predict the sum of six digits with $> 80\%$ accuracy, without having access to individual image labels.
Performance using set-pooling alone is similar.

\subsection{Point Cloud Classification}\label{sec:pointcloud}
A low-dimensional point-cloud is a set of low-dimensional vectors. This type of data is frequently 
encountered in various applications from robotics and vision to cosmology. 
In these applications, point-cloud data is often converted to voxel or mesh representation at a preprocessing step~\citep[\eg][]{maturana2015voxnet,ravanbakhshestimating,lin2004mesh}.
Since the output of many range sensors such as LiDAR -- which are extensively used in applications such as autonomous vehicles -- is in the form of point-cloud, direct application of deep learning methods to point-cloud is highly desirable. Moreover, when working with point-clouds rather than voxelized 3D objects, it is easy to apply transformations such as rotation and translation as differentiable layers at low cost.

\begin{table}
\caption{{\small \it Classification accuracy and the (size of) representation used by different methods on the ModelNet40 dataset.}}\label{table:point_cloud}
\scalebox{.78}{
 \begin{tabular}{||l | c c c ||} 
 \hline
 model & instance size &  representation & accuracy \\ [0.5ex] 
 \hline
  \textbf{set-layer} + transformation (ours) & \textbf{$\mathbf{5000 \times 3}$} &  point-cloud  &$90 \pm .3 \%$ \\ [0.5ex] 
  \textbf{set-layer} (ours) & \textbf{$\mathbf{1000 \times 3}$} &  point-cloud  &$87 \pm 1 \%$ \\ [0.5ex] 
  \textbf{set-pooling only} (ours) & \textbf{$\mathbf{1000 \times 3}$} &  point-cloud  &$83 \pm 1 \%$ \\ [0.5ex] 
  \textbf{set-layer} (ours) & \textbf{$\mathbf{100 \times 3}$} &  point-cloud  &$82 \pm 2 \%$ \\ [0.5ex] 
  {KNN graph-convolution} (ours) & $1000 \times (3 + 8)$ &  directed 8-regular graph &$58 \pm 2 \%$ \\ [0.5ex] 
  3DShapeNets \citep{wu20153d} & $30^3$ &  voxels (using convolutional deep belief net) &$77\%$ \\ [0.5ex] 
  DeepPano \citep{shi2015deeppano} & $64 \times 160$ &  panoramic image  (2D CNN + angle-pooling)  &$77.64\%$ \\ [0.5ex] 
  VoxNet \citep{maturana2015voxnet} & $32^3$ &  voxels  (voxels from point-cloud + 3D CNN)  &$83.10\%$ \\ [0.5ex] 
  MVCNN \citep{su2015multi} & $164 \times 164 \times 12$ &  multi-vew images  (2D CNN + view-pooling)  &$90.1\%$ \\ [0.5ex] 
  VRN Ensemble \citep{brock2016generative} & $32^3$ &  voxels  (3D CNN, variational autoencoder)  &$95.54\%$ \\ [0.5ex] 
  3D GAN \citep{wu2016learning} & $64^3$ &  voxels  (3D CNN, generative adversarial training)  &$83.3\%$ \\ [0.5ex] 
 \hline
\end{tabular}
}
\end{table}


Here, we show that treating the point-cloud data as a set, we can use the set-equivariant layer of \cref{eq:5} to classify 
point-cloud representation of a subset of ShapeNet objects~\citep{chang2015shapenet}, called ModelNet40~\citep{wu20153d}. This subset consists of 3D representation of 9,843
training and 2,468 test instances belonging to 40 classes of objects; see~\cref{fig:modelnet_dataset}. We produce point-clouds with  100, 1000 and 5000 particles each ($x,y,z$-coordinates) from the mesh representation of objects using 
the point-cloud-library's sampling routine~\citep{Rusu_ICRA2011_PCL}. Each set is normalized by the initial layer of the deep network to have zero mean (along individual axes) and unit (global) variance. Additionally we experiment with the K-nearest neighbor graph of each point-cloud and report the results using graph-convolution; see \cref{sec:point_cloud_details} for model details.

 \cref{table:point_cloud} compares our method against the competition.\footnote{The error-bar on our results is due to variations depending on the choice of particles during test time and it is estimated over three trials.}  Note that we achieve our best accuracy using $5000\times3$ dimensional representation of each object, which is much smaller than most other methods. All other techniques use either voxelization or multiple view of the 3D object for classification. Interestingly, variations of view/angle-pooling~\citep[\eg][]{su2015multi,shi2015deeppano} can be interpreted as set-pooling where the class-label is invariant to permutation of different views. The results also shows that using fully-connected layers with set-pooling alone (without max-normalization over the set)
works relatively well.

We see that reducing the number of particles to only 100, still produces comparatively good results. Using graph-convolution is computationally more challenging and produces inferior results in this setting. The results using 5000 particles is also invariant to small changes in scale and rotation around the $z$-axis; see \cref{sec:point_cloud_details} for details.

\begin{figure}[H]
  \centering
\includegraphics[width=1\textwidth,trim={5em 4em 5em 0},clip]{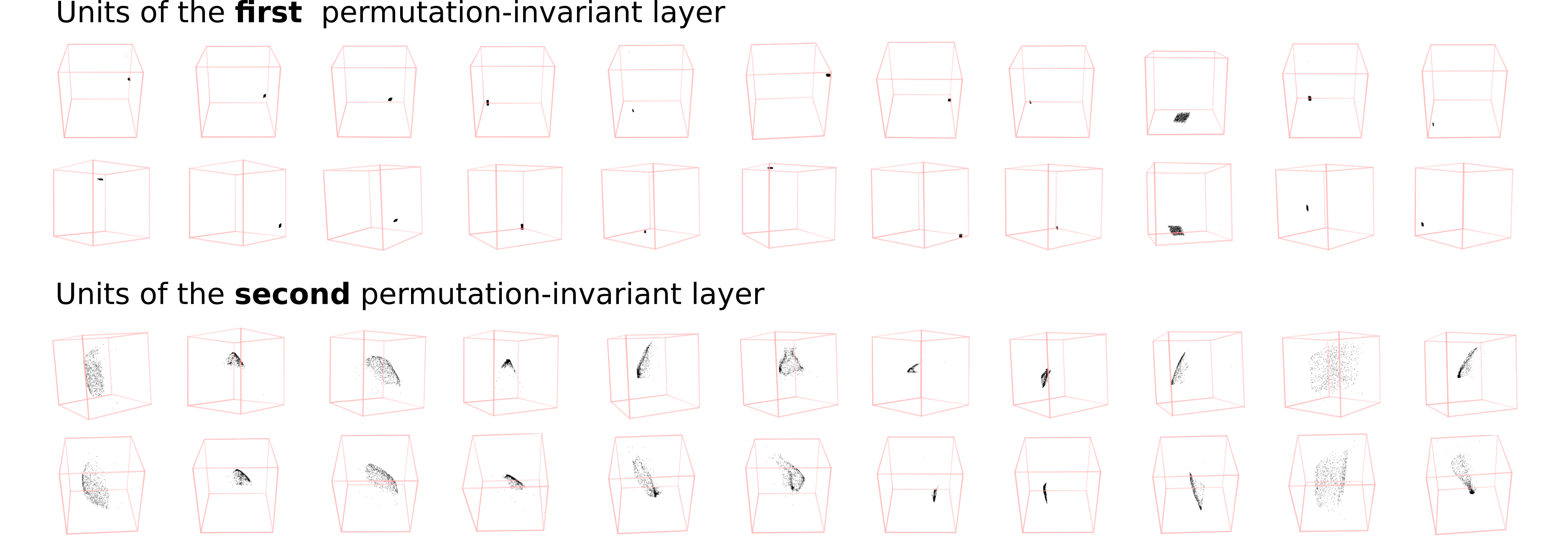} 
  \caption{\it \small  Each box is the particle-cloud maximizing the activation of a 
 unit at the firs (\textbf{top}) and second (\textbf{bottom}) permutation-equivariant layers of our model. Two images of the same column are two different views of the same point-cloud.} \label{fig:activations}
\end{figure}

\textbf{Features.}
To visualize the features learned by the set layers, 
we used Adamax~\citep{kingma2014adam} to locate 1000 particle coordinates maximizing the activation of each unit.\footnote{We started from uniformly distributed set of particles and used a learning rate of .01 for Adamax, with  first and second order moment of $.1$ and $.9$ respectively. We optimized the input in $10^5$ iterations. The results of \cref{fig:activations} are limited to instances where tanh units were successfully activated. Since the input at the first layer of our deep network is normalized to have a zero mean and unit standard deviation, we do not need to constrain the input while maximizing unit's activation.}
Activating the tanh units beyond the second layer proved to be difficult. 
\Cref{fig:activations} shows the particle-cloud-features
learned at the first and second layers of our deep network. We observed that the first layer
learns simple localized (often cubic) point-clouds at different $(x,y,z)$ locations, while the second layer learns more complex surfaces with different scales and orientations.

\section{Semi-Supervised Learning}\label{sec:semi_supervised}
In semi-supervised or transductive learning, some/all instances within each training set are labelled. Our goal is to 
make predictions for individual instances within a test set. Therefore, the permutation equivariant layer leverages
the interaction between the set-members to label individual member. Note that in this case, we do not perform any pooling operation
over the set dimension of the data.

\subsection{Set Anomaly Detection}\label{sec:celeb}
The objective here is for the deep model to find the anomalous face in each set, simply by observing examples and without any access to the attribute values. 
CelebA dataset~\citep{liu2015faceattributes} contains 202,599 face images, each annotated with 40 boolean attributes. We use $64\times64$ stamps and using these attributes we build 18,000 sets, each containing $N=16$ images (on the training set) as follows: after randomly selecting two attributes, we draw 15 images where those attributes are present
and a single image where both attributes are absent.  Using a similar procedure we build sets on the test images. No individual person's face appears in both train and test sets.

\begin{figure}[t]
  \centering
\includegraphics[width=\textwidth]{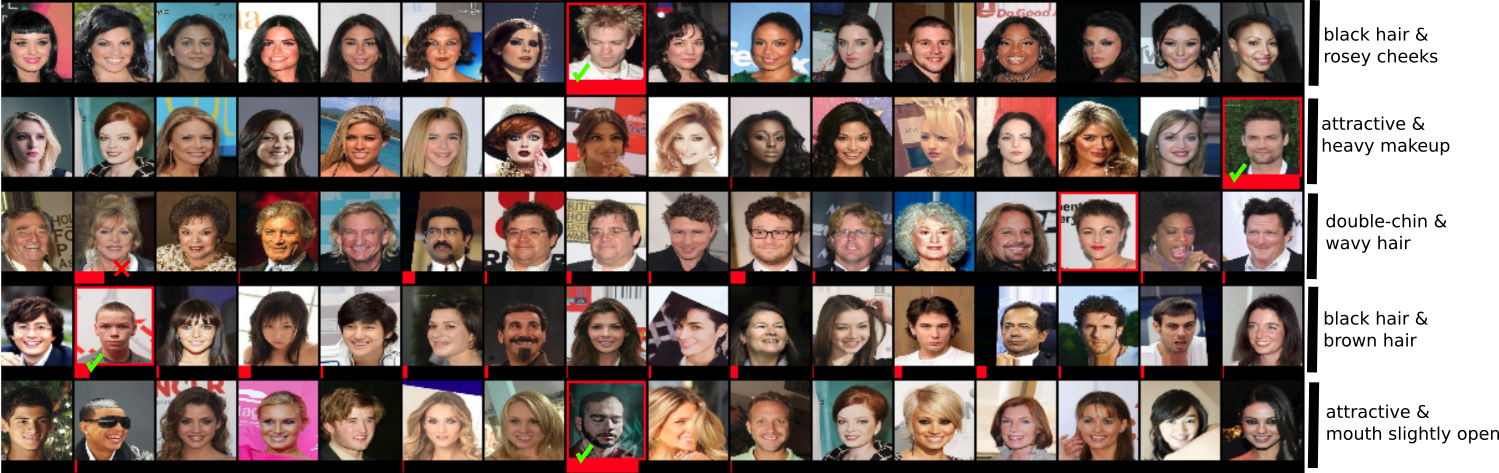} 
  \caption{\it \small Each row shows a set, constructed from CelebA dataset, such that all set members except for an outlier, share at least two attributes (on the right).
The \textbf{outlier is identified with a red frame}. The model is trained by observing examples of sets and their anomalous members, \textbf{without access to the attributes}.
The probability assigned to each member by the outlier detection network is visualized using a \textbf{red bar} at the bottom of each image. The probabilities in each row sum to one.
See \cref{sec:face_details} for more examples.
} \label{fig:faces}
\end{figure}

Our deep neural network consists of 9 2D-convolution and max-pooling layers followed by 3 permutation-equivariant layers and finally a softmax layer that assigns 
a probability value to each set member (Note that one could identify arbitrary number of outliers using a sigmoid activation at the output.)
Our trained model successfully finds the anomalous face in \textbf{75\% of test sets}. Visually inspecting these instances suggests that the task 
is non-trivial even for humans; see \cref{fig:faces}.
For details of the  model, training and more identification examples see \cref{sec:face_details}. 

As a \textit{baseline}, we repeat the same experiment by using a set-pooling layer after convolution layers, and replacing the permutation-equivariant layers with fully connected layers, with the same number of hidden units/output-channels, where the final layer is a 16-way softmax. The resulting network shares the convolution filters for all instances within all sets, however the input to the softmax
is not equivariant to the permutation of input images. Permutation equivariance seems to be crucial here as the baseline model achieves a training and \textbf{test accuracy of $\sim 6.3 \%$}; the same as random selection.

\section{Improved Red-shift Estimation Using Clustering Information}\label{sec:galaxy}
An important regression problem in cosmology is to estimate the red-shift of galaxies, corresponding to their age as well as their distance from us~\citep{binney1998galactic}.
Two common types of observation for distant galaxies include a) photometric and b) spectroscopic observations, where the latter can produce more accurate red-shift estimates.

One way to estimate the red-shift from photometric observations is using a regression model~\citep{connolly1995slicing}. We use a multi-layer Perceptron for this purpose
and use the more accurate spectroscopic red-shift estimates as the ground-truth.
As another baseline, we have a photometric redshift estimate that is provided by the catalogue and uses various observations (including clustering information)
to estimate individual galaxy-red-shift.
Our objective is to use clustering information of the galaxies to improve our red-shift prediction using the multi-layer Preceptron.

Note that the prediction for each galaxy does not change by permuting the members of the galaxy cluster. Therefore, 
we can treat each galaxy cluster as a ``set'' and use permutation-equivariant layer to 
estimate the individual galaxy red-shifts.
 
\begin{figure}[H]
  \centering
\includegraphics[width=\textwidth]{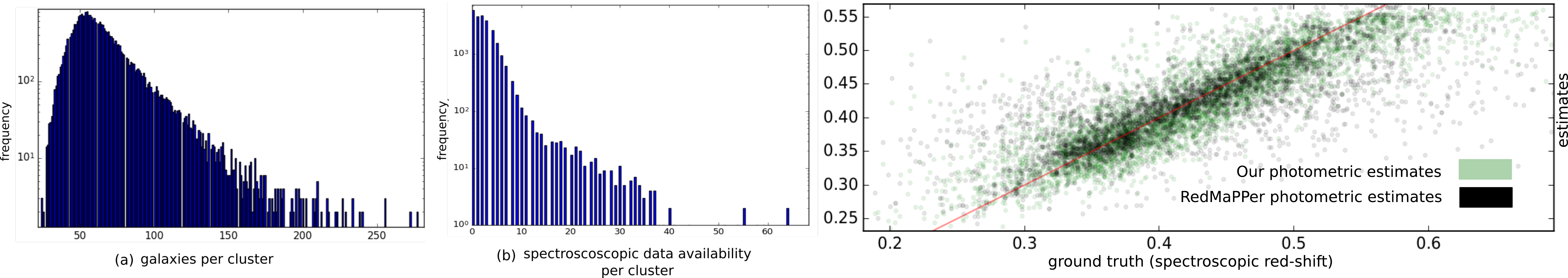} 
  \caption{\it \small 
  application of permutation-equivariant layer to semi-supervised red-shift prediction using clustering information:
  \textbf{a}) distribution of cluster (set) size; \textbf{b}) distribution of reliable red-shift estimates per cluster; \textbf{c}) prediction of red-shift on test-set (versus ground-truth) using clustering information as well as RedMaPPer photometric estimates (also using clustering information). 
}\label{fig:cluster_results}
\end{figure}

For each galaxy, we have $17$ photometric features \footnote{We have a single measurement for each u,g,r, i and z band as well as measurement error bars, location of the galaxy in the sky,  as well as the probability of each galaxy being the cluster center. We do not include the information regarding the richness estimates of the clusters from the catalog, for any of the methods, so that baseline multi-layer Preceptron is blind to the clusters.}
from the redMaPPer galaxy cluster catalog~\citep{rozo2014redmapper}, which contains photometric readings for 26,111 red galaxy clusters. 
In this task in contrast to the previous ones, sets have different cardinalities;
each galaxy-cluster in this catalog has between $\sim 20-300$ galaxies -- \ie $\xx \in \Re^{N(c) \times 17}$, where $N(c)$ is the cluster-size. See \cref{fig:cluster_results}(a) for distribution of cluster sizes. 
The catalog also provides accurate spectroscopic red-shift estimates for a \textit{subset} of these galaxies as well as photometric estimates that uses clustering information. 
\cref{fig:cluster_results}(b) reports the distribution of available spectroscopic red-shift estimates per cluster.

We randomly split the data into 90\% training and 10\% test clusters, and use the following simple architecture for semi-supervised learning.
We use four permutation-equivariant layers with 128, 128, 128 and 1 output channels respectively, where the output of the last layer is used as red-shift estimate. 
The squared loss of the prediction for available spectroscopic red-shifts is minimized.\footnote{We use mini-batches of size 128, Adam~\citep{kingma2014adam}, with learning rate of .001, $\beta_1=.9$ and $\beta_2=.999$.  
All layers except for the last layer use Tanh units and simultaneous dropout with 50\% dropout rate.}
\cref{fig:cluster_results}(c) shows the agreement of our estimates with spectroscopic readings on the galaxies in the test-set with spectroscopic readings.
The figure also compares the photometric estimates provided by the catalogue \citep[see][]{rozo2014redmapper}, to the ground-truth. 
As it is customary in cosmology literature, we report the average  \textbf{scatter} $\frac{|z_{\mathrm{spec}} - z|}{1 + z_{\mathrm{spec}}}$, where $z_{\mathrm{spec}}$ is the accurate spectroscopic measurement and $z$ is a photometric estimate.
The average scatter using \textbf{our model}  is $.023$ compared to the scatter of $.025$ in the \textbf{original photometric 
estimates} for the redMaPPer catalog. Both of these values are averaged over all the galaxies with spectroscopic measurements in the test-set.

We repeat this experiment, replacing the permutation-equivariant layers with fully connected layers (with the same number of parameters) 
and only use the individual galaxies with available spectroscopic estimate for training. The resulting average scatter for \textbf{multi-layer Perceptron} is $.026$,
demonstrating that using clustering information indeed improves photometric red-shift estimates.

\section*{Conclusion}
We introduced a simple parameter-sharing scheme to effectively achieve permutation-equivariance in deep networks and demonstrated
its effectiveness in several novel supervised and semi-supervised tasks.
Our treatment of set structure also generalizes various settings in multi-instance learning~\citep{ray2011multi,zhou2009multi}.
In addition to our experimental settings, the permutation-invariant layer 
can be used for distribution regression and classification which have become popular recently \citep{szabo16JMLR}. 
In our experiments with point-cloud data we observed the model to be robust to the variations in the number of
particles in each cloud, suggesting the usefulness of our method in the general setting of distribution regression -- where the number of 
samples should not qualitatively affect our representation of a distribution.
We leave further investigation of this direction to future work.

\section*{Acknowledgement}
We would like to thank Francois Lanusse for the pointing us to the redMaPPer dataset and 
the anonymous reviewers as well as Andrew Wagner for valuable feedback.

{\footnotesize{
\bibliography{iclr2017_conference}
\bibliographystyle{iclr2017_conference}
}
}
\clearpage
\appendix
{\Large Appendix}

\section{Proofs}\label{sec:proof}
\begin{proof} of the Theorem~\ref{theorem}\\
From definition of permutation equivariance $\ff_{\Theta}(\pi \xx) = \pi \ff_{\Theta}(\xx)$ and definition of $\ff$ in \cref{eq:1}, the condition becomes
 $\boldsymbol{\sigma}( \Theta \pi \xx) = \pi \boldsymbol{\sigma}( \Theta \xx)$, which (assuming sigmoid is a bijection) is equivalent to $\Theta \pi = \pi \Theta$. Therefore we need to show that the necessary and sufficient conditions for the matrix $\Theta \in \Re^{N \times N}$ to commute with all permutation matrices $\pi \in \mathcal{S}_n$ is given by \cref{eq:2}.
We prove this in both directions:
   \begin{itemize}
   \item To see why $\Theta = {\lambda} \mathbf{I} + {\gamma} \; (\mathbf{1} \mathbf{1}^{\mathsf{T}})$ commutes with any permutation matrix, first note that commutativity is linear -- that is
$$\Theta_1 \pi = \pi \Theta_1 \wedge \Theta_2 \pi = \pi \Theta_2 \quad \Rightarrow \quad (a \Theta_1 +  b \Theta_2) \pi = \pi (a \Theta_1 +  b \Theta_2).$$
Since both Identity matrix $\mathbf{I}$, and constant matrix $\mathbf{1} \mathbf{1}^\mathsf{T}$, commute with any permutation matrix, so does their linear combination $\Theta = {\lambda} \mathbf{I} + {\gamma} \; (\mathbf{1} \mathbf{1}^{\mathsf{T}})$.
\item We need to show that in a matrix $\Theta$ that commutes with ``all'' permutation matrices
  \begin{itemize}
  \item \textit{All diagonal elements are identical}: Let $\pi_{k,l}$ for $1\leq k,l\leq N, k \neq l$, be a transposition (\ie a permutation that only swaps two elements). The inverse permutation matrix of $\pi_{k,l}$ is the permutation matrix of $\pi_{l,k} = \pi_{k,l}^{\mathsf{T}}$. We see that commutativity of $\Theta$ with the transposition $\pi_{k,l}$ implies that $\Theta_{k,k} = \Theta_{l,l}$: 
    \begin{align*}
  \pi_{k,l} \Theta = \Theta \pi_{k,l} \; \Rightarrow  \; \pi_{k,l} \Theta \pi_{l,k} = \Theta \; \Rightarrow \;  (\pi_{k,l} \Theta \pi_{l,k})_{l,l} = \Theta_{l,l} \; \Rightarrow \;  \Theta_{k,k} = \Theta_{l,l}
    \end{align*}
Therefore, $\pi$ and $\Theta$ commute for any permutation $\pi$, they also commute for any transposition $\pi_{k,l}$ and therefore $\Theta_{i,i} = \lambda\, \forall i$.
\item \textit{All off-diagonal elements are identical}: We show that since $\Theta$ commutes with 
any product of transpositions, any choice two off-diagonal elements should be identical.
Let $(i,j)$ and $(i',j')$ be the index of two off-diagonal elements (\ie $i \neq j$ and $i' \neq j'$).
Moreover for now assume $i \neq i'$ and $j \neq j'$. Application of the transposition $\pi_{i,i'} \Theta$, swaps the rows $i,i'$ in $\Theta$.
Similarly, $\Theta \pi_{j,j'}$ switches the $j^{th}$ column with $j'^{th}$ column. From commutativity property of $\Theta$ and $\pi \in \mathcal{S}_n$ we have
\begin{align*}
 \pi_{j',j}\pi_{i,i'} \Theta =  \Theta \pi_{j',j}\pi_{i,i'} \; &\Rightarrow \pi_{j',j}\pi_{i,i'} \Theta (\pi_{j',j}\pi_{i,i'})^{-1} = \Theta \; &\Rightarrow \\
 \pi_{j',j}\pi_{i,i'} \Theta \pi_{i',i} \pi_{j, j'} = \Theta \; &\Rightarrow (\pi_{j',j}\pi_{i,i'} \Theta \pi_{i',i} \pi_{j, j'})_{i,j} = \Theta_{i,j} \; &\Rightarrow \; \Theta_{i',j'} = \Theta_{i,j}
\end{align*}
where in the last step we used our assumptions that $i \neq i'$, $j \neq j'$, $i \neq j$ and $i' \neq j'$. In the cases where either $i = i'$ or $j = j'$, we can use the above to show that $\Theta_{i,j} = \Theta_{i'', j''}$ and  $\Theta_{i',j'} = \Theta_{i'', j''}$, for some $i'' \neq i,i'$ and $j'' \neq j,j'$, and conclude $\Theta_{i,j} = \Theta_{i', j'}$.
  \end{itemize}
  \end{itemize}
\end{proof}

\section{Details of Models}\label{sec:details}
In the following, all our implementations use Tensorflow~\citep{abadi2016tensorflow}.

\subsection{MNIST Summation}\label{sec:mnist_details}
All nonlinearities are exponential linear units \citep[ELU][]{clevert2015fast}.
All models have 4 convolution layers followed by max-pooling. The convolution layers have 
respectively 16-32-64-128 output channels and $5\times5$ receptive fields.

Each pooling, fully-connected and set-layer is followed by a 20\% dropout. 
For models III and IV we use
simultaneous dropout.  
In models I and II, the convolution layers are followed by two fully-connected layers with 128 hidden units.
In model III, after the first fully connected layer we perform set-pooling followed by another dense layer with 
128 hidden units. In the model IV, the convolution layers are followed by a permutation-equivariant layer with 128
output channels, followed by set-pooling and a fully connected layer with 128 hidden units.
For optimization, we used a learning rate of .0003 with Adam using the default $\beta_1=.9$ and $\beta_2 = .999$.

\subsection{Face Outlier Detection Model}\label{sec:face_details}
Our model has 9 convolution layers with $3\times 3$ receptive fields. The model has convolution layers with 
$32,32,64$ feature-maps followed by max-pooling followed by 2D convolution layers with $64,64,128$ feature-maps 
followed by another max-pooling layer. The final set of convolution layers have $128,128,256$ feature-maps, 
followed by a max-pooling layer with pool-size of $5$ that reduces the output dimension to $\mathrm{batch-size} . N \times 256$, where the set-size $N = 16$.
This is then forwarded to three permutation-equivariant layers with $256, 128$ and $1$ output channels.
The output of final layer is fed to the Softmax, to identify the outlier. We use exponential linear units~\citep{clevert2015fast}, drop out with 20\% dropout rate 
at convolutional layers and 50\% dropout rate at the first two set layers. When applied to set layers, the selected feature (channel) is simultaneously
dropped in all the set members of that particular set. We use Adam~\citep{kingma2014adam} for optimization and use batch-normalization only in the convolutional layers.
We use mini-batches of $8$ sets, for a total of $128$ images per batch.

\begin{figure}
  \centering
\includegraphics[width=\textwidth]{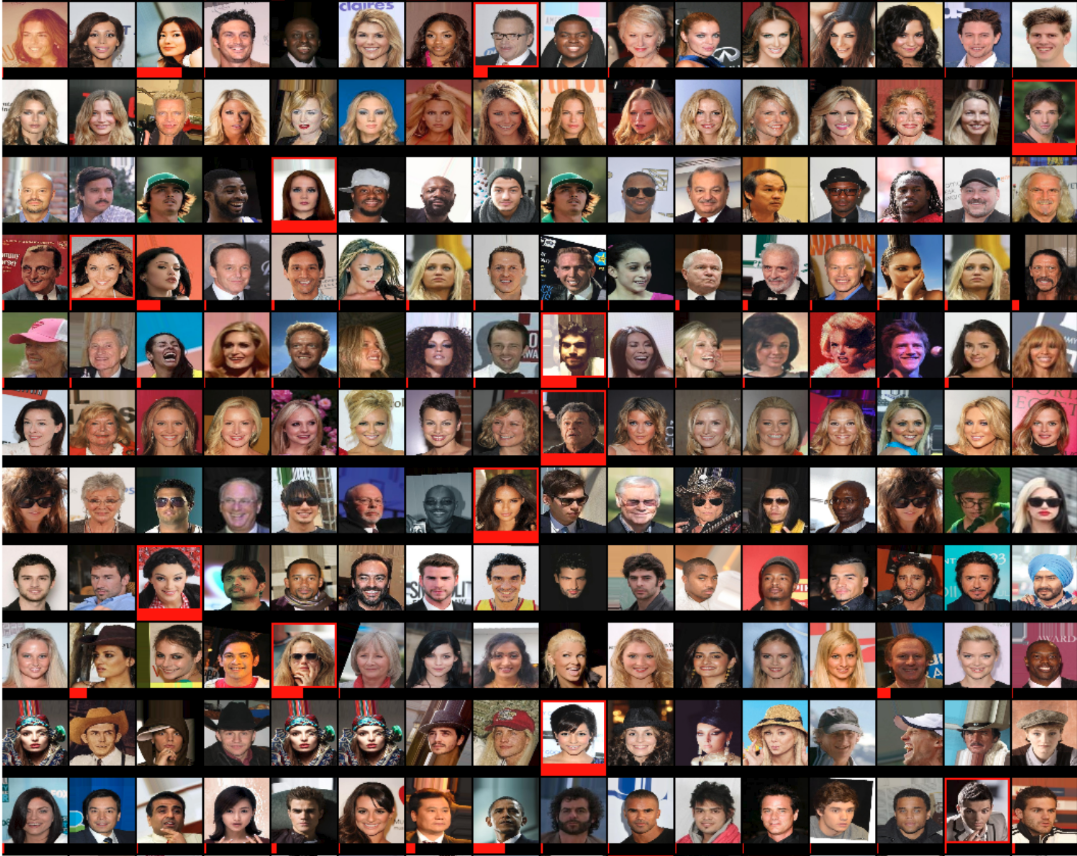} 
  \caption{\it \small Each row of the images shows a set, constructed from CelebA dataset images, such that all set members except for an outlier, share at least two attributes.
The \textbf{outlier is identified with a red frame}. The model is trained by observing examples of sets and their anomalous members and \textbf{without access to the attributes}.
The probability assigned to each member by the outlier detection network is visualized using a \textbf{red bar} at the bottom of each image. The probabilities in each row sum to one.
}\label{fig:more}
\end{figure}

\subsection{Models for Point-Clouds Classification}\label{sec:point_cloud_details}
\textbf{Set convolution.}
We use a network comprising of 3 permutation-equivariant layers with 256 channels followed by max-pooling over the set structure. The resulting vector representation of the set is then fed to a fully connected layer with 256 units followed by a 40-way softmax unit. We use Tanh activation at all layers and dropout on the layers after set-max-pooling (\ie two dropout operations) with 50\% dropout rate. Applying dropout to permutation-equivariant layers for point-cloud data deteriorated the performance. We observed that using different types of permutation-equivariant layers (see \cref{sec:parameter_sharing}) and as few as 64 channels for set layers changes the result by less than $5\%$ in classification accuracy.

For the setting with 5000 particles, we increase the number of units to 512 in all layers and randomly rotate the input around the $z$-axis. We also randomly scale the point-cloud by $s \sim \mathcal{U}(.8,1./.8)$. For this setting only, we use Adamax~\citep{kingma2014adam} instead of Adam and reduce learning rate from $.001$ to $.0005$.

\textbf{Graph convolution.} For each point-cloud instance with 1000 particles, we build a sparse K-nearest neighbor graph and use the three point coordinates as input features. We normalized all graphs at the preprocessing step. For direct comparison with set layer, we use the exact architecture of 3 graph-convolution layer followed by set-pooling (global graph pooling) and dense layer with 256 units.
We use exponential linear activation function instead of Tanh as it performs better for graphs. Due to over-fitting, we use a heavy dropout of 50\% after graph-convolution and dense layers. Similar to dropout for sets, all the randomly selected features are simultaneously dropped across the graph nodes. the  We use a mini-batch size of 64 and Adam for optimization where the learning rate is .001 (the same as that of permutation-equivariant counter-part).

Despite our efficient sparse implementation using Tensorflow, graph-convolution is significantly slower than the set layer. This prevented a thorough search for hyper-parameters and it is quite possible that better hyper-parameter tuning would improve the results that we report here.

\end{document}